\documentclass[conference]{IEEEtran}
\usepackage{amsmath,amssymb,amsfonts}
\usepackage{graphicx}
\usepackage{textcomp}
\usepackage{xcolor}
\def\BibTeX{{\rm B\kern-.05em{\sc i\kern-.025em b}\kern-.08em
    T\kern-.1667em\lower.7ex\hbox{E}\kern-.125emX}}
    
\usepackage{amsmath}
\usepackage{amssymb}
\usepackage{algorithm}
\usepackage{algorithmic}

\usepackage{epsfig}

\newtheorem{mydef}{Definition}[section]
\usepackage{booktabs} 
\usepackage{makecell} 

\usepackage{siunitx} 
\usepackage{url}
\usepackage{amsmath,amssymb,amsfonts}
\usepackage{graphicx}
\usepackage{textcomp}
\usepackage{adjustbox}
\usepackage{multirow}
\usepackage{makecell, longtable}
\usepackage{hhline}
\usepackage{subcaption} 
\usepackage{dsfont} 
    
\usepackage[square,numbers]{natbib}
\usepackage{url}
\usepackage[hidelinks]{hyperref} 
\hypersetup{
	colorlinks   = true, 
	urlcolor     = blue, 
	linkcolor    = blue, 
	citecolor   = red 
} 

\begin{document}

\title{Statistics and Deep Learning-based Hybrid Model for Interpretable Anomaly Detection\\
}

\author{\IEEEauthorblockN{Thabang Mathonsi}
\IEEEauthorblockA{\textit{School of Computer Science and Applied Mathematics} \\
\textit{University of the Witwatersrand}\\
Johannesburg, South Africa \\
thabang.mathonsi@wits.ac.za}
\and
\IEEEauthorblockN{Terence L van Zyl}
\IEEEauthorblockA{\textit{Institute for Intelligent Systems} \\
\textit{University of Johannesburg}, South Africa \\
tvanzyl@uj.ac.za}
}

\maketitle

\begin{abstract}
Hybrid methods have been shown to outperform pure statistical and pure deep learning methods at both forecasting tasks, and at quantifying the uncertainty associated with those forecasts (prediction intervals). One example is Multivariate Exponential Smoothing Long Short-Term Memory (MES-LSTM), a hybrid between a multivariate statistical forecasting model and a Recurrent Neural Network variant, Long Short-Term Memory. It has also been shown that a model that ($i$) produces accurate forecasts and ($ii$) is able to quantify the associated predictive uncertainty satisfactorily, can be successfully adapted to a model suitable for anomaly detection tasks. With the increasing ubiquity of multivariate data, and new application domains, there have been numerous anomaly detection methods proposed in recent years. The proposed methods have largely focused on deep learning techniques, which are prone to suffer from challenges such as ($i$) large sets of parameters that may be computationally intensive to tune, $(ii)$ returning too many false positives rendering the techniques impractical for use, $(iii)$ requiring labeled datasets for training which are often not prevalent in real life, and ($iv$) understanding of the root causes of anomaly occurrences inhibited by the predominantly black-box nature of deep learning methods. In this article, an extension of MES-LSTM is presented, an interpretable anomaly detection model that overcomes these challenges. With a focus on renewable energy generation as an application domain, the proposed approach is benchmarked against the state-of-the-art. The findings are that MES-LSTM anomaly detector is at least competitive to the benchmarks at anomaly detection tasks, and less prone to learning from spurious effects than the benchmarks, thus making it more reliable at root cause discovery and explanation.

\end{abstract}


\section{Introduction}

Time series anomaly detection is useful in applications from a variety of industries~\cite{Ala19}. In their analytical study comparing classical and deep learning-based anomaly detection models, \citet{muni19} observe that deep learning outperforms the classical approaches. In another study, \citet{Mat2} conclude that deep learning is at least competitive to statistical-based models. However, there is still a relative gap that exists for hybrid approaches in anomaly detection within the multivariate setting. Furthermore, root cause discovery and explainability remains an open research problem in the context of multivariate anomaly detection~\cite{rad21, jac21}.

\citet{Mat3} present a statistics and deep learning-based hybrid model, Multivariate Exponential Smoothing Long Short-Term Memory (MES-LSTM). The method leverages multivariate exponential smoothing within the pre- and postprocessing modules. It also circumvents some problems associated with large sets of parameters requiring tuning, computational cost at training, and extensive inference time. These problems are circumvented by incorporating a small parsimonious recurrent deep neural network for learning, for instance, the cross-dependency structure within the covariates.

In this paper, MES-LSTM is extended to the task of anomaly detection. We also investigate the potential of explainability and interpretability for such a model. These goals are executed with an application to the renewable energy domain.

\subsection{Literarure Review}
\label{sub:litreview_fusion}

The literature review is segmented into $(i)$ work that has been presented in the research community relating to explainable anomaly detection, and $(ii)$ explainer systems or techniques for interpreting anomaly detection models, explanation discovery, and root cause analysis.

Furthermore, a video can also be considered time series when the streaming images are taken as a matrix of pixel values with coordinates and a time dimension. As such, some techniques that were traditionally applied to streaming video have also been adapted and extended to fit time series anomaly detection problems. As such, techniques from computer vision have not been excluded in the review of review recent advances and the state-of-the-art.

\subsubsection{Explainable Anomaly Detection}
Some work has been done in unsupervised machine learning. \citet{nguy19} for instance, consider the problem of anomaly detection in networks data, with an application to an Internet Service Provider example. The authors show their approach, using variational autoencoders, is able to effectively detect malicious attacks on a network. They use the gradients from the autoencoders for model interpretability.

\citet{rad21} consider the problem of explainable anomaly detection framed within a high dimensionality setting. The authors report competitive model performance without significant gains in computational cost.

\citet{carl19} offer a technique for interpreting Isolation Forests (IF)~\cite{liu08}, a commonly used model for anomaly detection tasks. They limit their focus to the scenario of Industry 4.0., with a particular focus on root cause analysis. Root cause analysis, as an application domain for explainable anomaly detection, emphasizes the need for robust models that are deployable in realistic industrial settings~\cite{jeya19, ma20}.

\citet{west21} consider explainable anomaly detection within a framework for procurement fraud identification. To this end, they consider real data from a government department in Singapore. The authors report their techniques, in real-life deployment, resulted in cost and time reduction of up to 10\% compared to previously applied compliance checks. There lack in ubiquity of such studies illustrates the need for models that do not suffer from high computational cost. Great computational expense is one of the biggest criticisms of deep learning as it hinders real-time deployment in real-world applications~\cite{Bri16}.

\citet{liz21} use an explainable convolutional model for one class image classification, and detail some challenges from spurious effects such as watermarks. A similar approach is followed by \citet{pang21}. A distinguishing factor is that the latter considers both training with no anomalies, and trainingwith anomalous observations as well. Considering the problem of streaming images, \citet{wu21} apply denoising autoencoders to surveillance video. The authors report competitive results with reduced computational time.

In terms of image classification, \citet{ruff21} offer a comprehensive review of other techniques that have been applied in the field, and systematically compare their performance on benchmark examples. Another notable review~\cite{pang20} looks at deep anomaly detection, and critically analyses the advantages and disadvantages of various techniques. Others, such as the works of \citet{chal19} and \citet{kira17}, only consider different classes of models applied to specific application domains. The interested reader is also directed to a discussion of explainability in health data~\cite{tjoa20}, financial data~\cite{guid19}, and object detection and recognition in image data~\cite{selv17, mitc19}. For a more general discussion of concepts related to explainability such as interpretability, and understandability, there are targeted resources such as \citet{arri20}.

Applications for time series classification or anomaly detection can be discerned from the plurality of public dataset repositories, such as the UCR~\cite{dau19ucr} or the UEA archive~\cite{bagn18uea}, for instance. Such applications, with a focus on explainable anomaly detection, include predictive maintenance at industrial sites~\cite{sera21}, or rule extraction for unsupervised anomaly detection conducted~\cite{barb22}.

Some models have been successful at classification and anomaly detection tasks, but due to their complex hierarchical architectures, incorporating explainability proves difficult.  Some of these techniques are discussed here for completeness. One such advancement is an ensemble method, Collective of Transformation-based Ensembles (COTE)~\cite{bagn16}, where 35 models are ensembled over different time series representations based on transformations such as time warping~\cite{kate16} or shapelets~\cite{bost17}, for instance.

COTE was further extended by \citet{line18}, using Hierarchical Voting (HIVE-COTE). HIVE-COTE performed well over the UCR and UEA archives~\cite{bagn17}. Using a weighted probabilistic ensemble~\cite{larg19}, this ensemble approach combines Shapelet Transform Classifier (STC)~\cite{hill14}, Contractable Bag of Symbolic-Fourier Approximation Symbols (CBOSS)~\cite{midd19}, Time Series Forest (TSF)~\cite{Deng13} and Random Interval Spectral Ensemble (RISE)~\cite{line18}.

\subsubsection{Explainability Systems and Methods}

When categorized based on scope of the explanation, explainer systems offer either local or global explanations. Local explanations explain a single prediction result over the entire model, i.e., it explains the conditional interaction between dependent and independent variables with respect to the single prediction. As mentioned by \citet{ribe16}, the explanation are required to make sense within a local setting. In the context of the current study, this means that one explanation should be valid in some region encompassing immediate \emph{neighbors}. The immediate neighbors are understood to be anomalous observations occurring around the same time, and of the same type.

Explainability of anomalies can also be conducted for a (potentially large) \emph{set} of anomalies, for example, in the form of rule lists. These are called global explanations. Finding a truly global explanation, one that applies to multiple anomalous observations of different types occurring at different times is a difficult task~\cite{jac21}. As such, global explanation are usually aggregates of different explanations, or the most representative explanations for the entire model. 

Another distinction can be drawn between model-specific and model-agnostic explainer systems. The former are applicable to certain kinds of model(s) (say for instance, strictly convolutional or strictly recurrent models, but usually not applicable to both) while the latter can be applied to multiple models.

Yet another distinction can be drawn between feature attribution, path attribution, and association rule mining techniques. Feature attribution determines the contribution of each feature towards the model’s prediction for a given input example. This attribution shows the relationship between a feature and the prediction. As a result, users are able to understand which features their network relies on.

Path attribution methods explain the output of the model that is based on gradients. That means the contribution of each feature is computed by aggregating the gradients from baseline values to the current input along the path. One such method is Path Integrated Gradients (PIG)~\cite{sund17}.

In contrast, association rule mining finds correlations and co-occurrences between features in a large dataset. They are considered as most interpretable prediction models with their simple if-then rules. A rule is essentially an if-then statement with two components: an antecedent and a consequent. The input feature with a condition is an antecedent and a prediction is its consequent. The popular techniques to extract the rules from a large dataset are Scalable Bayesian Rule Lists (SBRL)~\cite{hong17}, Gini Regularization (GiniReg)~\cite{burk20} and Rule Regularization (RuleReg)~\cite{burk19}. Such techniques have been applied successfully to classifiers in surveillance tasks~\cite{veer21}.

\citet{arri20} discuss transparent models that automatically incorporate explainability such as Logistic regression, decision trees, and nearest neighbour models, as well as post-hoc models, that are explainable with the aid of an additional technique.

Explainability techniques that have been developed for or are typically used for image-based models include Deep Learning Important FeaTures (DeepLIFT)~\cite{shri17}, Local Explanation Method using Nonlinear Approximation (LEMNA)~\cite{guo18}, and Gradient-weighted Class Activation Mapping (Grad-CAM)~\cite{selv17}. These explainer systems usually output heatmaps~\cite{lapu19} or saliency visualisations~\cite{simo13} that rank the feature importance of input images input to the network. A good example of saliency maps is presented by \citet{sidd19}, for example, with application to convolutional layers.

For time series models, techniques often employed are Model Agnostic Supervised Local Explanations (MAPLE)~\cite{plum18}, Local Interpretable Model-agnostic Explanations (LIME)~\cite{ribe16}, Local rule-based explanations (LORE)~\cite{guid18}, Learning to explain (L2X)~\cite{chen18} and Shapley additive explanations (SHAP)~\cite{lund17}. As a cautionary note in particular for time series modeling, there is difficulty due to temporal dependence inherent in the data. As a consequence, surrogate solutions such as LIME or SHAP neglect the chronological sequence ordering of the model inputs.

LIME~\cite{ribe16} explains model inferences by using a local interpretable sparse linear model as an approximation. Anchors~\cite{Ribe18} offers an incremental improvement on LIME by replacing the linear model used as proxy with a logical rule for explaining inferences. Anchors offers better coverage and explainability of anomalous neighbors, but is not readily applicable to time series data. Other local explainer systems that rank feature importance include responsibility scores (RESP)~\cite{bert20} and axiomatic attribution~\cite{sund17}.

\subsection{Motivation}
It is straightforward to motivate for deep learning as a mechanism for solving time series classification problems and anomaly detection tasks. One reason is to leverage the feature learning abilities of deep learning algorithms~\cite{neam18}. Deep neural networks have also performed well at other tasks requiring temporal sequence modeling (similar to time series) such as natural language processing~\cite{bahd15} and speech recognition~\cite{sain13}. Lastly, deep learning has been shown to scale well with increased dimensionality~\cite{keog17}.

However, there exists some challenges with the deep learning approach. These include large sets of parameters that may be computationally expensive to tune, and long inference time that may be impractical in settings that require fast reliable information as feedback from models~\cite{Mat3}.

As evidenced from existing scholarly research, there is a great need for explanation discovery and interpretable anomaly detection with real-world applications such as root cause analysis~\cite{liu08, jeya19, ma20}. There is a need to circumvent the computational cost and time complexity usually associated with deep learning that prevents them from being used outside of a laboratory setting, and enables deployment in real-world applications such as in the compliance study conducted by \citet{west21} or the streaming video study by \citet{wu21}.

In addition, learning from spurious effects can contaminate the root cause and explanation discovery leading to stakeholders making bad decisions informed by incorrectly explained model inferences. It is important to minimize the effects of learning from, say, random noise in time series data or even watermarks in image data~\cite{liz21}.

As a final point, this study may be motivated using another factor from real world applicability. If an anomaly is identified, it might be time consuming for a domain expert or human agent to inspect all the components that may have possibly contributed to the anomalous event in order to identify the root cause. It may be more useful to the inspector if, for instance, a model is able to narrow the search space down to a reasonable fraction of components that are most probable to have contributed to the anomaly.

\subsection{Contribution}
The novel contribution can be summarized as follows:
\begin{itemize}
	\item a statistics and deep learning-based forecast machinery is extended to anomaly detection tasks;
	\item this new hybrid anomaly detection model ($i$) incorporates a dynamic threshold-setting approach, which learns and adapts the applicable threshold as new information becomes available, and ($ii$) functions within a semi-supervised framework, so no golden labels are required for training or detecting the thresholds for detecting anomalies; and
	\item the presented approach is augmented with explainability and interpretability, thus enabling root cause analysis, and how well the model avoids learning from spurious effects using a novel metric.
\end{itemize}

\section{Methodology}

Renewable energy resources, such as wind/solar farms, affect the grid in a different way compared to conventional fossil fuel generators due to their stochastic nature. In particular, the uncertain disturbances introduced by renewables may compromise operational grid safety. This scenario emphasizes the need for system operators to accurately identify disturbances in a timeous fashion. They are then able to perform corrective measures timely so as to ensure the safety of the grid.

System operators have access to streaming time-stamped measurements, from monitors such as phasor measurement units. These measurements enable system operators to answer critical questions including $(i)$ When is an event happening? $(ii)$ What type of event is happening? and $(iii)$ Where is the source that caused the event? These are the research questions stated succinctly, and they would be phrased equivalently for other domains besides renewable energy regeneration.

Following the methodology of \citet{Zhen21}, who first proposed the Power Systems Machine Learning dataset  (PSML)~\cite{psml} for use within the machine learning for decarbonized energy grids domain, the problem can be formulated as follows.

\subsection{Problem Statement}
The streaming measurements can be denoted by $X\in \mathbb{R}^{N\times K}$, where $N$ is the number of available observations and $K$ is the number of measurements or covariates.

\emph{Event detection} aims to answer the first question above, by identifying an oscillation occurrence when or if it happens. Answering this question involves using a model $\mathcal{H}$ to identify the oscillation occurrence given sequence $X$, i.e., $\mathcal{H}: X\rightarrow \{0, 1\}$. Suppose an event occurs at time $t_{*}$: an alarm should be raised when $t_m^ \geq t^*$, and as soon as possible ($1$ predicted).

\emph{Event classification} answers the second question above based on streaming sensor measurements. Given the observations $X$, the model $\mathcal{H}$ must classify the underlying event type $\xi$, i.e., 
$\mathcal{H}: X \rightarrow \xi$. PSML presents a truly multivariate problem as $\xi$ is more than just binary classification (i.e. normal or anomalous), but it constitutes a subset of disturbances $\mathcal{C}$ where
$\mathcal{C}:=$ $\{$branch fault, branch tripping, bus fault, bus tripping, generator tripping, forced oscillation$\}$. This problem framing emphasizes the need to keep track of multiple streams of data with interdependent covariates that are autocorrelated interacting within the global grid. By observing variables such as voltage from each bus in the system, the aim is to determine based on thresholds, for example, if and what kind of event is occurring.

\emph{Event localization} focuses on locating events from disturbances $\mathcal{C}$, or the root cause of events (for forced oscillations) by observing measurements. The model $\mathcal{H}$ must map measurements $X$ to the bus(es) $z$ nearest to the events detected or the root cause of the events, i.e., $\mathcal{H}: X\rightarrow z$, where $z$ is a subset of buses $\mathcal{Z}$ in the entire system.


\subsection{Algorithms}

The following benchmark models are used: InceptionTime~\cite{fawa20}, multi-channels deep convolutional neural networks (MC-DCNN)~\cite{zhen14}, Residual Network (ResNet)~\cite{wang17}, Time series attentional prototype network (TapNet)~\cite{zhan20}, Minimal random convolutional kernel transform (MiniRocket)~\cite{dem21}, one-Nearest neighbour with Euclidean distance (NNEuclid)~\cite{line15}, independent dynamic time warping (iDTW)~\cite{shok17}, and dependent dynamic time warping (dDTW)~\cite{shok17}. The architectures of the different benchmark models are briefly described next.

\subsubsection{Classical Models}

\paragraph{Nearest Neighbour}
The first of classical model is one-Nearest neighbour with Euclidean distance (NNEuclid). Nearest neighbour classifiers with a distance function have been among the most popular techniques for time series classification~\cite{line15}. In one study, classifiers with dynamic time warping distance perform well as baselines~\cite{bagn17}. In another, \citet{line15} shows dynamic time warping is at least competitive to all the other distance measures considered. Interestingly, the best performers in the study are reported to be ensembling neural network classifiers combined with different distance measures.

\paragraph{Dynamic Time Warping}
Dynamic Time Warping (DTW) can be applied to time series data composed of varying length, but for simplicity, the following description is limited to the case involving series of equal length, such as presented by \citet{bagn17}. The distance between two equal length series $\boldsymbol{a} = (a_1, a_2, \, \dots, a_m)$ and $\boldsymbol{b} = (b_1, b_2, \, \dots, b_m)$ is calculated as follows:
\begin{enumerate}
    \item $\boldsymbol{\psi}$ is a matrix sized $m \times m$ where $\boldsymbol{\psi}_{i,\, r} = (a_i - b_r)^2$
    \item A warping path $\rho = ((e_1, f_1), (e_2, f_2), \dots, (e_s, f_s))$ is a contiguous set of matrix indices from $\boldsymbol{\psi}$, subject the constraints:
    \begin{itemize}
        \item $(e_1, f_1) = (1, 1)$
        \item $(e_s, f_s) = (m, m)$
        \item $0 \leq e_{i+1} - e_i \leq 1 \, \forall \, i < m$
        \item $0 \leq f_{i+1} - f_i \leq 1 \, \forall \, i < m$
    \end{itemize}
    \item Let $p_i = \boldsymbol{\psi}_{e_i ,f_i}$, then the path distance is $\mathcal{D}_p = \sum_{i = 1}^{m} p_i$
    \item Multiple warping paths exists, but the aim is to find a path that minimizes the accumulative distance $\rho^{*} = \min_{p \in \rho} \mathcal{D}_p(\boldsymbol{a}, \boldsymbol{b})$
    \item Solving the following relation yields the optimal distance:
    \begin{align}
        \text{DTW}_{(i, r)} &= \boldsymbol{\psi}_{i, r} + \min \begin{cases}
    \text{DTW}_{(i - 1, r)} \\
    \text{DTW}_{(i, r - 1)} \\
    \text{DTW}_{(i - 1, r - 1)}
  \end{cases},
    \end{align}
    where the final distance is given by $\text{DTW}_{(m, m)}$.
\end{enumerate}

Some improvements may be applied to DTW for increased efficiency, such as constraining deviations from the diagonal but this falls beyond the cope of this paper. \citet{shok17} defines strategies for applying DTW to multivariate setting. These are independent and dependent approaches.

The independent method, iDTW, as the name suggests, has a separate treatment for each dimension. Using a separate distance matrix for each dimension, iDTW then sums the resulting time warping distances:
\begin{align}
    \text{iDTW}_{i, r} (\boldsymbol{x}_{\boldsymbol{a}}, \boldsymbol{x}_{\boldsymbol{b}}) &= \sum_{k =  1}^d \text{DTW}(x_{a, i, k} - x_{b, r, k})^2
\end{align}

The main idea behind Dependent dynamic time warping (dDTW) is the assumption that the accurate warping is identical for all the dimensions. Given a single time series, the matrix $\boldsymbol{\psi}_{i,r}$ is no longer considered the distance between two points, but is redefined as the Euclidean distance between the two vectors that constitute a representation of the full dimensional space. The dependant strategy is more efficient as warping is simultaneous for all the dimensions, and the distance between steps $i$ and $r$ in terms of time resolution is given by:
\begin{align}
    \boldsymbol{\psi}_{i, r} (\boldsymbol{x}_{\boldsymbol{u}}, \boldsymbol{x}_{\boldsymbol{v}}) &= \sum_{k =  1}^d (\boldsymbol{x}_{\boldsymbol{u, k}}, \boldsymbol{x}_{\boldsymbol{v, k}})^2.
\end{align}

There also exists an adaptive strategy~\cite{shok17} for selecting between independent and dependent dynamic time warping. How the distance is chosen depends on an instance-by-instance threshold deducible from the training data. Adaptive time warping falls beyond the scope of the current study.


\subsubsection{Deep Learning-based Models}
\paragraph{MiniRocket}
MiniRocket~\cite{dem21} is adapted from Rocket~\cite{demp20} which was ranked among the best performers on multiple datasets in a recent study~\cite{ruiz21}. The authors report MiniRocket is at most 75 times faster that Rocket, with comparative accuracy. Rocket combines convolution kernels that are randomly initialised, with a linear classifier, typically ridge regression or logistic regression. The method produces feature maps where the maximum value the proportion of positive values (ppv) are extracted.

Hyperparameter tuning is restricted to the following search spaces. The length $\varsigma \in \left\{7, 9, 11\right\}$; the kernel weights $w_i \sim \mathcal{N}(0, 1)$; the dilation, $d$, is sampled from the exponential distribution; and whether or not the series is padded is decided with equal probability.

In contrast, MiniRocket attempts to minimize the randomness characteristic of Rocket. It achieves this by pre-assigning values to a subset of the hyperparameters discussed above, or limiting the search space to a smaller grid, yet still reportedly achieving comparable accuracy. The changes are summarized in Table~\ref{tab:minirocket}~\cite{dem21}, where $\mathcal{N}$ is the normal distribution and $\mathcal{U}$ is the uniform distribution.

\begin{table}[h]
\centering
\caption{Summary of changes from Rocket\cite{demp20} to MiniRocket~\cite{dem21}.}
\label{tab:minirocket}
\resizebox{1\columnwidth}{!}{%
\begin{tabular}{@{}lll@{}}
\toprule
& \textbf{Rocket}      & \textbf{MiniRocket}     \\ \midrule \midrule
length             & $\lbrace{7, 9, 11\rbrace}$         & 9                       \\
weights            & $\mathcal{N}(0, 1)$  & $\lbrace{-1, 2\rbrace}$               \\
bias               & $\mathcal{U}(-1, 1)$ & from convolution output \\
dilation           & random               & fixed                   \\
padding            & random               & fixed                   \\
features           & ppv, max             & ppv                     \\
number of features & 20,000                & 10,000                   \\ \bottomrule
\end{tabular}}
\end{table}

\paragraph{MC-DCNN}
Multi Channel Deep Convolutional Neural Network (MC-DCNN)~\cite{zhen14} is a modification of conventional deep convolutional neural networks. The convolutions are applied independently per covariate in the multivariate input space.

Every dimension of the multivariate input data goes through two convolutional stages with eight filters each of length five and configured with ReLU activation functions. After each convolution there is a max-pooling operation, followed by a fully connected layer. Softmax is used for the final classification.

\paragraph{ResNet}
ResNet~\cite{wang17} architecture has three convolutional layers within each of three residual blocks, followed by a Global Average Pooling (GAP) layer. The main idea behind ResNet is the use of residual shortcuts connecting consecutive convolutional layers. The key difference when compared with conventional convolutions from fully convolutional networks for instance, is the addition of these linear shortcuts. The shortcuts reduce the vanishing gradient effect~\cite{he16}, by enabling the gradient to flow directly through these connections. In a recent study~\cite{ruiz21}, ResNet ranked among the best performers on multiple datasets.

\paragraph{InceptionTime}
InceptionTime incorporates ResNet~\cite{wang17} and Inception modules~\cite{szeg15}. An Inception module takes as input multivariate series of size $m \times k$. By using a bottleneck layer with length and stride one, it reduces the dimensionality to $m \times k$ where $k^{\prime} < k$. InceptionTime assigns random initial weights to five instances of the artificial neural network and ensembles them  for greater stability~\cite{fawa20}. One out of the five networks replaces the three blocks of the aforementioned three classical convolutional layers from ResNet with two blocks of three Inception modules each. However, the new blocks also maintain residual connections, and they too are followed by GAP and softmax layers.

\paragraph{TapNet}
The final benchmark model considered combines classical and deep learning-based traits. \citet{zhan20} note that deep learning methods are good at learning low dimensional features and classical approaches such as dynamic time warping are competitive for applications involving small datasets. TapNet, combining these traits, has a network architecture composed of three modules: Random Dimension Permutation, Multivariate Time Series Encoding and Attentional Prototype Learning. These modules can further be broken down into fully connected layers, three sets of convolutional layers that are one dimension each, a global pooling layer, batch normalisation, and Leaky Rectified Linear Units (LReLU)~\cite{Maas13}.

\subsection{Anomaly Detection}

The benchmark models and their hyperparameters are kept constant from the original manuscripts that the techniques were initially introduced. For MES-LSTM, similarly, the model architecture as described by \citet{Mat3} is retained. This hybrid model is used in conjunction with the methodology presented by \citet{Mat2}, in particular Algorithm~\ref{algo:ad} is employed in order to adapt the forecast machinery for the task of anomaly detection.

\begin{algorithm}[h]
\caption{Anomaly Detection}\label{algo:ad}

\begin{algorithmic}
\IF{$U_t \leq y_t \leq L_t$}
  \STATE $y_t$ is normal
\ELSE
  \IF{$\text{IS}_\alpha \left( y_t \right) \geq 1.33 \times \text{IS}_\alpha \left( y_* \right)$ \AND $ y_t > 10 \times \text{std} {\left\lbrace \dots, y_{t-3}, y_{t-2}, y_{t-1} \right\rbrace}$}
  \STATE $y_t$ is anomalous
  \COMMENT{where $y_*$ is the last anomalous observation}
  \ENDIF
\ENDIF
\end{algorithmic}
\end{algorithm}

Training time series are extracted from the millisecond transient phasor measurement units data. Training samples are randomly selected amounting to 439 time series, and the remaining 110 time series are used for testing (20\%). Each sequence has metadata associated with the event type similar to the classification use case, i.e. \emph{branch fault, branch trip, bus fault, bus trip, gen trip}. Each time series has a sequence length of 960 observations, representing 4s in the system recorded at 240Hz. There are 91 dimensions for each time series, including voltage, current and power measurements across the transmission system. The experiment is repeated 35 times to mitigate the stochastic nature of the deep learning models.

\subsection{Interpretability}
The presented approach uses model-agnostic feature attribution techniques. LIME~\cite{ribe16} is a local explainability technique suitable for local explanations. This method perturbs the input in the neighbourhood of an instance and examines the output of the model. LIME, thus, indicates the input features the model considers when making a prediction. LIME works by using a proxy based on a simple model, intrinsically interpretable, such as a linear regression model. The surrogate model is applied around each prediction between the input variable space and the corresponding outcome variables space. Explainability is discerned from the main anomaly detection model by perturbing the input variables of a multivariate observation and tracking how the predictions change.

SHAP~\cite{lund17} use Shapley values from cooperative game theory, which indicates what reward players can expect depending on a coalition function. To extend this approach to the explainability of artificial intelligence agents, players are considered as features and reward as the outcome of the model. Although SHAP also provides global interpretability (behaviour of the entire model), this paper focuses on its ability to shed light on local interpretability (behaviour of a single prediction). This local focus enables the use of SHAP in conjunction with LIME and facilitates comparison. The importance rank of feature $i$ is deduced by taking all subsets of features except feature $i$, $D \setminus \boldsymbol{x_i}$, and computing the effect of the output predictions after adding feature $i$ back to all the subsets previously extracted. All the contributions are then combined to compute the marginal contribution of the each feature.

The labeled dataset used in this study stipulates which predictor contributed most to an anomaly. Ideally, for the interpretability component to be useful for stakeholders, the correct contributor should be identified and ranked high up in terms of feature importance. To ensure this, a novel metric is presented, that can be tuned to the appropriate task-specific sensitivity.

\begin{mydef}[Mean Discovery Score]
    Let $\beta$ indicate the task-specific sensitivity, i.e. ideally, the principal contributing predictor $\kappa$ should be ranked in the highest $\beta$ features in terms of importance. Let the $i^{\textnormal{th}}$ out-of-sample observation that is in fact anomalous be denoted by $y_i \in \left\{ a \right\}$, where $\left\{ a \right\}$ is the set of all anomalies. Then, by aggregating how many times this ranking occurs at the specific sensitivity, the mean discovery score (MDS$_{\beta}$) is given by
    \begin{align}
        \text{MDS}_{\beta} &= \dfrac{1}{\mathbf{card}\left(\left\{ a \right\}^{A_i}_D \right)} \sum_{t = 1}^{m} \mathbb{1}_{{y_i \in \left\{ a \right\}} \cap {\left\lbrace \mathcal{R}(\kappa_i) \leq \beta \right\rbrace}},
    \end{align}
    where the rank of an attribute's feature importance is denoted by $\mathcal{R}$, and $\mathbf{card}\left(\left\{ a \right\}^{A_i}_D \right)$ is how many times an algorithm $A_i$ is able to detect anomalies in dataset $D$.
\end{mydef}

MDS$_{\beta}$ is useful for scoring the explainability of accurate anomaly detection models, and would not be suitable for use if the set of anomalies correctly detected $\left\{ a \right\}$ by algorithm $A_i$ is small in comparison to the set of overall anomalous events.

\subsection{Analysis}

InceptionTime, MC-DCNN and ResNet are implemented in Tensorflow, TapNet and MiniRocket in Pytorch, and the DTW techniques from scratch. The code implementations in this study retain the default settings detailed by the respective authors in their original manuscripts. Some algorithms have modules embedded that perform hyperparameter tuning. In cases where this is applicable, the hyperparameter optimization modules are kept as is, but no additional tuning is conducted. Below, the configurations for each algorithm is detailed.

For DTW the full warping window is used. MiniRocket is configured with a ridge regression classifier and 10,000 kernels. TapNet uses defaults set to 500 trees, 3,000 epochs, a learning rate of $10^{-4}$, weight decay of $10^{-2}$, stop threshold of $10^{-8}$, number of filters given by 256, 256, and 128 respectively, kernels by 8, 5, and 3 respectively, while dilation is one with no dropout. ResNet has 1,500 epochs, a batch size of 16, learns at a rate of $10^{-2}$ which, if no improvement is observed for 50 epochs, is set to $5^{-2}$. ResNet is configured with three residual blocks composed of three convolutional layers each, where the sizes of the kernels are 8, 5, and 3 respectively, and 64, 128, and 128 filters respectively per convolutional layer for each block. InceptionTime runs for 1,500 epochs with a batch size of 64. InceptionTime is configured with dual residual blocks, each composed of three Inception modules where the sizes of the kernels are sizes 10, 20, and 40 respectively per module, and learns at a rate of $10^{-2}$, which, if no improvement is observed for 50 epochs, is set to $5^{-2}$. 

\section{Results and Discussion}
This section analyzes the results of the performance of the proposed method compared to the state of the art. Both the anomaly detection and the interpretability tasks are analyzed and discussed.

\subsection{Anomaly Detection}

Table~\ref{tab:au_ROC} details the aggregated results for the area under the ROC (auROC) curve. When observing the standard deviation, the deterministic models all have no variability in their results, i.e. 1NN, and the dynamic time warping models. InceptionTime shows the most variability and this property may be somewhat undesirable within the context of assisting domain experts. A good anomaly detection model should have results that are not only accurate, but are also consistent over multiple trials. In this regard of variability, ResNet has the most consistent results from the non-deterministic models.

Table~\ref{tab:au_ROC} also indicates that MC-DCNN is the best performing anomaly detection model, followed closely by MES-LSTM, ResNet, InceptionTime and MiniRocket. TapNet is not much better than the deterministic distance-based models.

\begin{table*}[htbp]
\centering
		\caption{Area Under Receiver Operating Characteristic curve for all Models over all Trials.}
		\label{tab:au_ROC}
		\resizebox{1\textwidth}{!}{%
\begin{tabular}{@{}lrrrrrrrrr@{}}
\toprule
 &
  \textbf{1NN} &
  \textbf{dDTW} &
  \textbf{iDTW} &
  \textbf{InceptionTime} &
  \textbf{MC-DCNN} &
  \textbf{MES-LSTM} &
  \textbf{MiniRocket} &
  \textbf{ResNet} &
  \textbf{TapNet} \\ \midrule \midrule
\textbf{mean} &
  0.3301 &
  0.4146 &
  0.4500 &
  0.5822 &
  0.7547 &
  0.7376 &
  0.5675 &
  0.7012 &
  0.4804 \\
\textbf{std} &
  0.0000 &
  0.0000 &
  0.0000 &
  0.1025 &
  0.0500 &
  0.0736 &
  0.0925 &
  0.0358 &
  0.0812 \\ \bottomrule
\end{tabular}
}
\end{table*}

\begin{figure}[h!tbp]
	\centering 
	\includegraphics[width=\columnwidth]{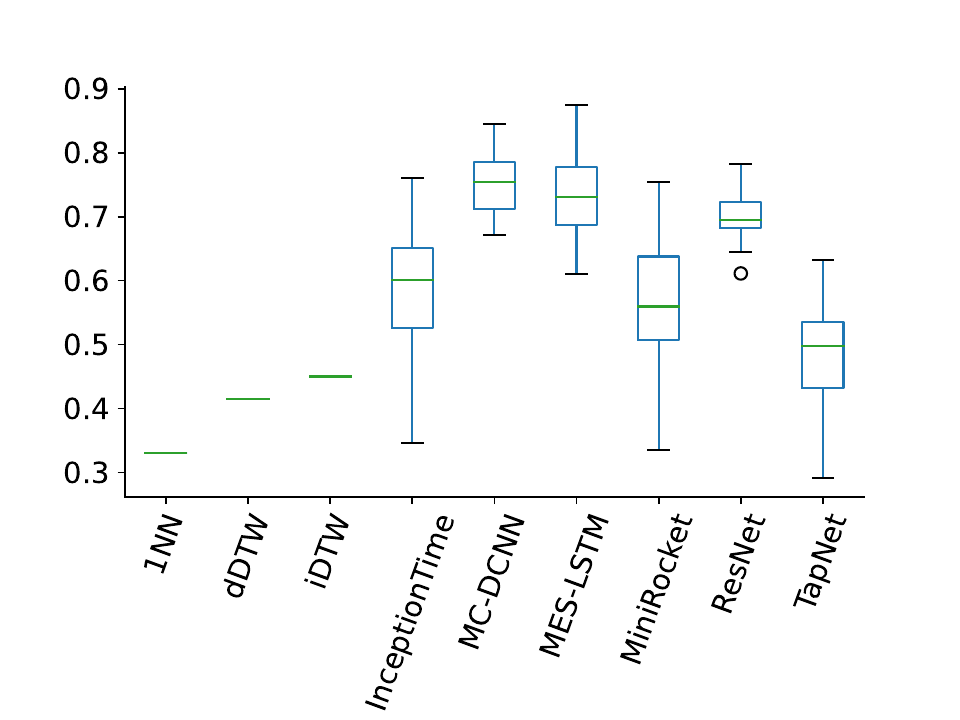}
	\caption{Area Under the ROC curve Distribution Boxplot for all Trials.\label{fig:roc_box}}
\end{figure}

The box and whisker plot in Figure~\ref{fig:roc_box} indicate MES-LSTM achieves the highest performance score on a single trial (highest whisker end-point), although MC-DCNN has the highest overall mean aggregated over all trials. Inception Time and MiniRocket have the highest variability in terms of performance results, whilst ResNet is the most consistent model.

Examining the area under the Precision-Recall (auPR) curve in Table~\ref{tab:au_PR} reaffirms the above discussion. The main difference is in the range of performance scores. The range is higher across all the models as the auPR metric takes into account the class imbalance inherent in the data.
\begin{table*}[htbp]
\centering
		\caption{Area Under Precision-Recall curve for all Models over all Trials.}
		\label{tab:au_PR}
		\resizebox{1\textwidth}{!}{%
\begin{tabular}{@{}lrrrrrrrrr@{}}
\toprule
 &
  \textbf{1NN} &
  \textbf{dDTW} &
  \textbf{iDTW} &
  \textbf{InceptionTime} &
  \textbf{MC-DCNN} &
  \textbf{MES-LSTM} &
  \textbf{MiniRocket} &
  \textbf{ResNet} &
  \textbf{TapNet} \\ \midrule \midrule
\textbf{mean} &
  0.4225 &
  0.4803 &
  0.5500 &
  0.7332 &
  0.8470 &
  0.8421 &
  0.6359 &
  0.8117 &
  0.5604 \\
\textbf{std} &
  0.0000 &
  0.0000 &
  0.0000 &
  0.1087 &
  0.0328 &
  0.0549 &
  0.0594 &
  0.0170 &
  0.0471 \\ \bottomrule
\end{tabular}

}
\end{table*}\vspace{-9pt}

\begin{figure}[htbp]
	\centering 
	\includegraphics[width=\columnwidth]{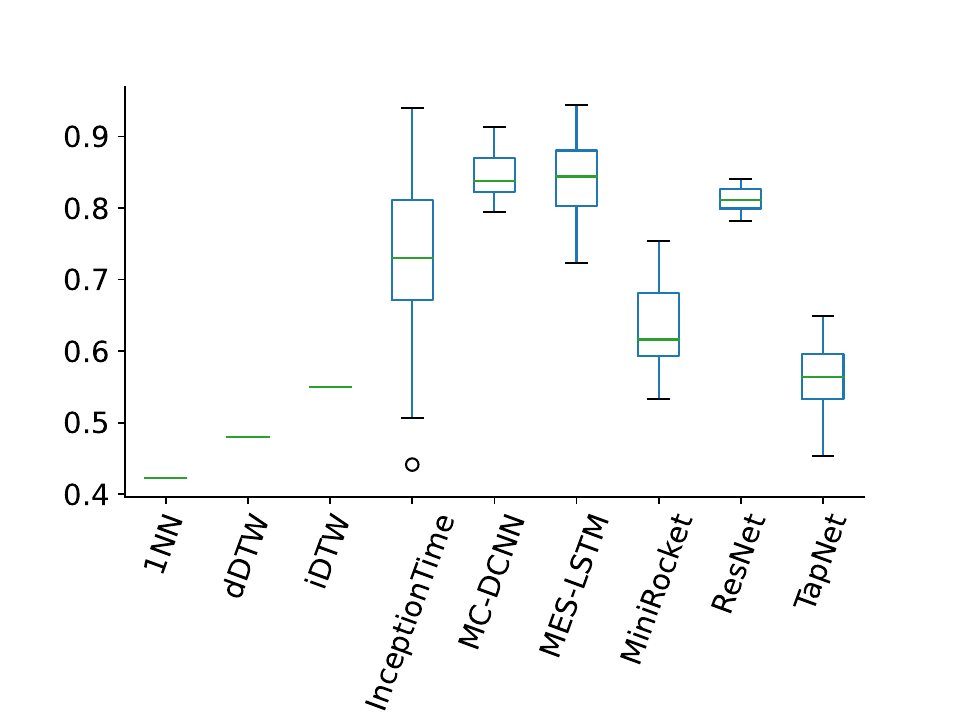}
	\caption{Area Under the PR curve Distribution Boxplot for all Trials.\label{fig:pr_box}}
\end{figure}

The box and whisker plot in Figure~\ref{fig:pr_box} further shows the maximum auPR is achieved by InceptionTime, at the cost of the aforementioned variability (which also adversely impacts the overall performance mean). With regards to highest overall performance mean, the top five models are, in order from best, MC-DCNN, MES-LSTM, ResNet, InceptionTime, and MiniRocket.

A Student's t-test for statistical significance is conducted at the $\alpha = 0.01$ level of significance, for the performance results in terms of anomaly detection. The null hypothesis is H$_0$: the benchmark models outperform MES-LSTM. From Tables~\ref{tab:test_ROC} and Table~\ref{tab:test_PR}, the only instance where one is unable to reject the null hypothesis at the $\alpha = 0.01$ level of significance is for MC-DCNN.

\begin{table*}[h!tbp]
\centering
		\caption{Student's t-test for Testing Significance of auROC Performance Results (H$_0$: Benchmark
Models Outperform MES-LSTM).}
		\label{tab:test_ROC}
		\resizebox{1\textwidth}{!}{%
\begin{tabular}{@{}lrrrrrrrr@{}}
\toprule
 &
  \multicolumn{1}{l}{\textbf{1NN}} &
  \multicolumn{1}{l}{\textbf{dDTW}} &
  \multicolumn{1}{l}{\textbf{iDTW}} &
  \multicolumn{1}{l}{\textbf{InceptionTime}} &
  \multicolumn{1}{l}{\textbf{MC-DCNN}} &
  \multicolumn{1}{l}{\textbf{MiniRocket}} &
  \multicolumn{1}{l}{\textbf{ResNet}} &
  \multicolumn{1}{l}{\textbf{TapNet}} \\ \midrule \midrule
\textbf{statistic} &
  32.7568 &
  25.9649 &
  23.1195 &
  7.2838 &
  -1.1331 &
  8.5118 &
  2.6362 &
  13.8820 \\
\textbf{p-value} &
  0.0000 &
  0.0000 &
  0.0000 &
  0.0000 &
  0.8692 &
  0.0000 &
  0.0056 &
  0.0000 \\ \bottomrule
\end{tabular}

}
\end{table*}\vspace{-9pt}

\begin{table*}[htbp]
\centering
		\caption{Student's t-test for Testing Significance of auPR Performance Results (H$_0$: Benchmark
Models Outperform MES-LSTM).}
		\label{tab:test_PR}
		\resizebox{1\textwidth}{!}{%
\begin{tabular}{@{}lrrrrrrrr@{}}
\toprule
 &
  \textbf{1NN} &
  \textbf{dDTW} &
  \textbf{iDTW} &
  \textbf{InceptionTime} &
  \textbf{MC-DCNN} &
  \textbf{MiniRocket} &
  \textbf{ResNet} &
  \textbf{TapNet} \\ \midrule \midrule
\textbf{statistic} &
  45.2511 &
  39.0184 &
  31.5026 &
  5.2917 &
  -0.4495 &
  15.0871 &
  3.1367 &
  23.0563 \\
\textbf{p-value} &
  0.0000 &
  0.0000 &
  0.0000 &
  0.0000 &
  0.6726 &
  0.0000 &
  0.0016 &
  0.0000 \\ \bottomrule
\end{tabular}
}
\end{table*}\vspace{-9pt}

\subsection{Interpretabilty and Explainability}
Since there are 96 covariates in total, a starting point would be to consider what a domain expert might consider useful inference from a machine learning tool. In case of power trip, for instance, it would be time consuming to check all 96 possible contributors. However, checking a reasonable subset would be more feasible. Below, $\beta$ is set to elements in the range $\left\{5, 10, 15\right\}$, although this range can be determined by requirements specific to a particular use-case scenario. The rationale behind the chosen range is ideally, a good explainer system should rank the chief contributor in the first five highest ranked features (saving the domain expert the most amount of time); an explainer with moderate skill would rank the chief contributor in the five to first ten highest ranked features, while the worst would rank the chief contributor in the first ten to 15 highest ranked features and beyond.

Below, only the top four performing models are considered. Anything that offers less than 50\% in accuracy for anomaly detection can be argued to be no better than random guessing. TapNet, although above 50\% in performance score, does not report anomaly detection performance skill much higher than the distance-based methods and is also not included in the discussion that follows.

Table~\ref{tab:mds_lime} shows MES-LSTM has the highest correct attribution at $\beta = 5$ features considered, followed by MC-DCNN, InceptionTime and ResNet. At $\beta = 10$ and at $\beta = 15$ features considered, MES-LSTM and InceptionTime are in the top two. ResNet peaks at around 80\% correct attribution at $\beta = 15$ features considered, while InceptionTime has the highest overall score at 94.61\%. Not one of the models reaches 100\%, indicating that there are still some missing key contributing factors not accounted for even after 15 covariates are explored in terms of feature importance.  
From Table~\ref{tab:mds_shap} it is deducible that at $\beta = 5$ features considered, at most only 73\% explainability is accounted for. The maximum score is achieved by MES-LSTM at $\beta = 15$ features considered.

\begin{table}[htbp]
\centering
		\caption{Mean Discovery Score for LIME applied to Top Four Anomaly Detectors.}
		\label{tab:mds_lime}
		\resizebox{\columnwidth}{!}{%
			\begin{tabular}{@{}lrrrr@{}}
\toprule
\textbf{$\beta$} & \textbf{InceptionTime} & \textbf{MC-DCNN} & \textbf{MES-LSTM} & \textbf{ResNet} \\ \midrule \midrule
5       & 0.7113        & 0.7554    & 0.7634     & 0.6419   \\
10      & 0.9059        & 0.8222    & 0.8856     & 0.7025   \\
15      & 0.9461        & 0.8985    & 0.9346     & 0.8077   \\ \bottomrule
\end{tabular}
}
	\end{table}
	
\begin{table}[htbp]
\centering
		\caption{Mean Discovery Score for SHAP applied to Top Four Anomaly Detectors.}
		\label{tab:mds_shap}
		\resizebox{\columnwidth}{!}{%
\begin{tabular}{@{}lrrrr@{}}
\toprule
\textbf{$\beta$} & \textbf{InceptionTime} & \textbf{MC-DCNN} & \textbf{MES-LSTM} & \textbf{ResNet} \\ \midrule \midrule
\textbf{5}  & 0.7311 & 0.6871 & 0.7376 & 0.6437 \\
\textbf{10} & 0.8570 & 0.7380 & 0.9179 & 0.7205 \\
\textbf{15} & 0.9263 & 0.8219 & 0.9481 & 0.7754 \\ \bottomrule
\end{tabular}

}
\end{table}
	
\begin{figure*}[htbp]
	
	\resizebox{\textwidth}{!}{
		{\captionsetup{position=bottom,justification=centering}\begin{subfigure}{0.49\textwidth}
			\includegraphics[width=\columnwidth]{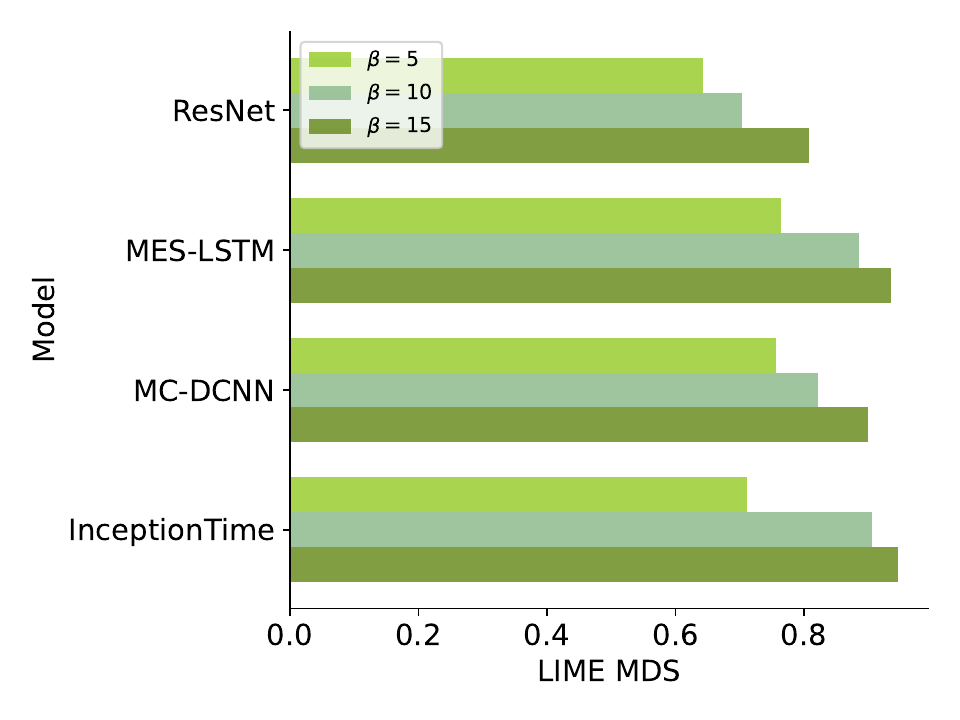}
			\caption{}
			\label{fig:mds_lime}
		\end{subfigure}
		\begin{subfigure}{0.49\textwidth}
			\includegraphics[width=\columnwidth]{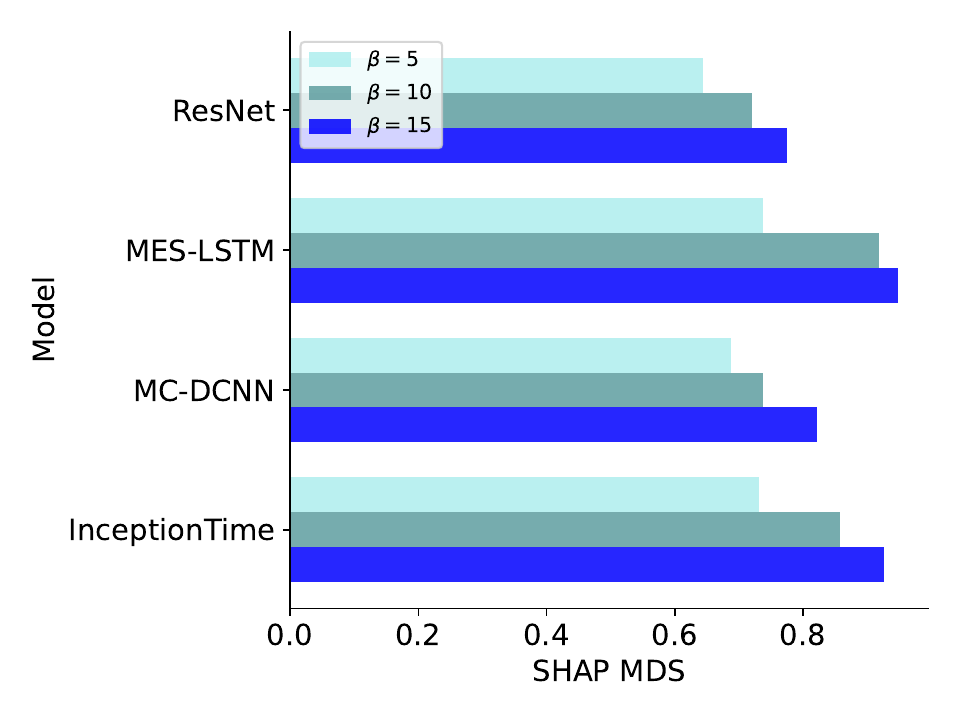}
			\caption{}
			\label{fig:mds_shap}
	\end{subfigure}}}
	\caption{Mean Discovery Score for Explainer Techniques applied to Top Four Anomaly Detectors. (\textbf{A}) LIME. (\textbf{B}) SHAP. \label{fig:mds}}
\end{figure*}

The discovery scores are illustrated graphically in Figure~\ref{fig:mds}. MES-LSTM has the highest correct attribution at all levels of features considered for both LIME and SHAP, except for LIME at $\beta = 10$ (where MES-LSTM is outperformed by InceptionTime).

\section{Conclusion}

There are two main objectives in this paper. One is to build an anomaly detection machinery that is a hybrid composed of statistical and deep learning techniques. The second is to incorporate interpretability to such a technique. The desired outcomes related to these objectives are one, that the anomaly detection skill is at least competitive to the state-of-the-art; and two, that the explainability component has a good level of correct attribution.

The proposed method is outperformed in some instances with regards to anomaly detection by, for example, MC-DCNN. MC-DCNN is able to model the spatial correlations well, and this could be a contributing factor to the superior performance. MES-LSTM is outperformed marginally and although competitive with regards to anomaly detection, the overall performance is an area for improvement.

However, when it comes to correctly attributing important features for the anomalies detected by each of the top four performing models, MES-LSTM is the overall highest achiever. The high discovery scores are as a result of the architecture's good modeling of temporal dependence. Accurate attribution is important as it ensures the model is not learning from spurious effects. It also reinforces trust for manual inspectors of, say, mechanical systems when an anomaly within the system is detected and possible causes are reported.

The voltage measurements and current measurements are governed by both time evolution from external oscillation events and as well as spatial dependency from the inherent network connectivity over the entire grid. The problem framing is of multivariate spatio-temporal anomaly detection and interpretability. Future work may involve graph neural networks, which show significant promise in the tasks underpinned by similar settings, such as in climate modeling~\cite{cach21}.

It is possible that through additional engineering of the benchmark algorithms and tuning of their hyperparameters, better overall performance could have been realized. However, the idea was to test the anomaly detectors based on the configuration recommendations suggested by the respective authors in their original manuscripts. The approach using original configurations mitigates biases that may result from optimising algorithms for particular datasets and particular tasks.

In this study, a novel metric is proposed for assessing usability of a model with regards to usefulness of the model's interpretability to a domain expert. There are many metrics for tasks such as forecasting and anomaly detection, but research centered around explainability is lacking in terms of metrics for ease of comparison among multiple models. Adding more metrics to measure the level of correct attribution is also an avenue for future research.

\bibliography{references}

\begin{thebibliography}{74}
\providecommand{\natexlab}[1]{#1}
\providecommand{\url}[1]{\texttt{#1}}
\expandafter\ifx\csname urlstyle\endcsname\relax
  \providecommand{\doi}[1]{doi: #1}\else
  \providecommand{\doi}{doi: \begingroup \urlstyle{rm}\Url}\fi

\bibitem[Alla~S.(2019)]{Ala19}
Adari~S.K. Alla~S.
\newblock \emph{Practical Use Cases of Anomaly Detection.}
\newblock Apress, Berkeley, CA., 2019.
\newblock \doi{10.1007/978-1-4842-5177-5_8}.

\bibitem[Bagnall et~al.(2016)Bagnall, Lines, Hills, and Bostrom]{bagn16}
Anthony Bagnall, Jason Lines, Jon Hills, and Aaron Bostrom.
\newblock Time-series classification with cote: The collective of
  transformation-based ensembles.
\newblock In \emph{2016 IEEE 32nd International Conference on Data Engineering
  (ICDE)}, pages 1548--1549, 2016.
\newblock \doi{10.1109/ICDE.2016.7498418}.

\bibitem[Bagnall et~al.(2017)Bagnall, Lines, Bostrom, Large, and Keogh]{bagn17}
Anthony Bagnall, Jason Lines, Aaron Bostrom, James Large, and Eamonn Keogh.
\newblock The great time series classification bake off: a review and
  experimental evaluation of recent algorithmic advances.
\newblock \emph{Data mining and knowledge discovery}, 31\penalty0 (3):\penalty0
  606--660, 2017.
\newblock \doi{10.1007/s10618-016-0483-9}.

\bibitem[Bagnall et~al.(2018)Bagnall, Dau, Lines, Flynn, Large, Bostrom,
  Southam, and Keogh]{bagn18uea}
Anthony~J. Bagnall, Hoang~Anh Dau, Jason Lines, Michael Flynn, James Large,
  Aaron Bostrom, Paul Southam, and Eamonn~J. Keogh.
\newblock The {UEA} multivariate time series classification archive, 2018.
\newblock \emph{CoRR}, abs/1811.00075, 2018.
\newblock URL \url{http://arxiv.org/abs/1811.00075}.

\bibitem[Bahdanau et~al.(2015)Bahdanau, Cho, and Bengio]{bahd15}
Dzmitry Bahdanau, Kyunghyun Cho, and Yoshua Bengio.
\newblock Neural machine translation by jointly learning to align and
  translate.
\newblock In Yoshua Bengio and Yann LeCun, editors, \emph{International
  Conference on Learning Representations}, 2015.
\newblock URL \url{http://arxiv.org/abs/1409.0473}.

\bibitem[Barbado et~al.(2022)Barbado, Corcho, and Benjamins]{barb22}
Alberto Barbado, Oscar Corcho, and Richard Benjamins.
\newblock Rule extraction in unsupervised anomaly detection for model
  explainability: Application to oneclass svm.
\newblock \emph{Expert Systems with Applications}, 189:\penalty0 116100, 2022.
\newblock ISSN 0957-4174.
\newblock \doi{10.1016/j.eswa.2021.116100}.

\bibitem[{Barredo Arrieta} et~al.(2020){Barredo Arrieta}, Díaz-Rodríguez,
  {Del Ser}, Bennetot, Tabik, Barbado, Garcia, Gil-Lopez, Molina, Benjamins,
  Chatila, and Herrera]{arri20}
Alejandro {Barredo Arrieta}, Natalia Díaz-Rodríguez, Javier {Del Ser}, Adrien
  Bennetot, Siham Tabik, Alberto Barbado, Salvador Garcia, Sergio Gil-Lopez,
  Daniel Molina, Richard Benjamins, Raja Chatila, and Francisco Herrera.
\newblock Explainable artificial intelligence (xai): Concepts, taxonomies,
  opportunities and challenges toward responsible ai.
\newblock \emph{Information Fusion}, 58:\penalty0 82--115, 2020.
\newblock ISSN 1566-2535.
\newblock \doi{10.1016/j.inffus.2019.12.012}.

\bibitem[Bertossi et~al.(2020)Bertossi, Li, Schleich, Suciu, and
  Vagena]{bert20}
Leopoldo Bertossi, Jordan Li, Maximilian Schleich, Dan Suciu, and Zografoula
  Vagena.
\newblock Causality-based explanation of classification outcomes.
\newblock In \emph{Proceedings of the Fourth International Workshop on Data
  Management for End-to-End Machine Learning}, DEEM'20, New York, NY, USA,
  2020. Association for Computing Machinery.
\newblock ISBN 9781450380232.
\newblock \doi{10.1145/3399579.3399865}.

\bibitem[Bostrom and Bagnall(2017)]{bost17}
Aaron Bostrom and Anthony Bagnall.
\newblock Binary shapelet transform for multiclass time series classification.
\newblock In \emph{Transactions on Large-Scale Data-and Knowledge-Centered
  Systems {XXXII}}, pages 24--46. Springer, 2017.
\newblock \doi{10.1007/978-3-662-55608-5_2}.

\bibitem[Brink et~al.(2016)Brink, Richards, and Fetherolf]{Bri16}
Henrik Brink, Joseph Richards, and Mark Fetherolf.
\newblock Scaling machine-learning workflows.
\newblock In \emph{Real-World Machine Learning}, chapter~9. Manning
  Publications Co., New York, NY, 2016.
\newblock URL
  \url{https://livebook.manning.com/book/real-world-machine-learning/chapter-9/8}.

\bibitem[Burkart et~al.(2019)Burkart, Huber, and Faller]{burk19}
Nadia Burkart, Marco Huber, and Phillip Faller.
\newblock Forcing interpretability for deep neural networks through rule-based
  regularization.
\newblock In \emph{2019 18th IEEE International Conference On Machine Learning
  And Applications (ICMLA)}, pages 700--705, 2019.
\newblock \doi{10.1109/ICMLA.2019.00126}.

\bibitem[Burkart et~al.(2020)Burkart, Faller, Peinsipp, and Huber]{burk20}
Nadia Burkart, Philipp~M. Faller, Elisabeth Peinsipp, and Marco~F. Huber.
\newblock Batch-wise regularization of deep neural networks for
  interpretability.
\newblock In \emph{2020 IEEE International Conference on Multisensor Fusion and
  Integration for Intelligent Systems (MFI)}, pages 216--222, 2020.
\newblock \doi{10.1109/MFI49285.2020.9235209}.

\bibitem[Cachay et~al.(2021)Cachay, Erickson, Bucker, Pokropek, Potosnak, Bire,
  Osei, and Lütjens]{cach21}
Salva~Rühling Cachay, Emma Erickson, Arthur Fender~C. Bucker, Ernest Pokropek,
  Willa Potosnak, Suyash Bire, Salomey Osei, and Björn Lütjens.
\newblock The world as a graph: Improving el {Ni\~no} forecasts with graph
  neural networks, 2021.

\bibitem[Carletti et~al.(2019)Carletti, Masiero, Beghi, and Susto]{carl19}
Mattia Carletti, Chiara Masiero, Alessandro Beghi, and Gian~Antonio Susto.
\newblock Explainable machine learning in industry 4.0: Evaluating feature
  importance in anomaly detection to enable root cause analysis.
\newblock In \emph{2019 IEEE International Conference on Systems, Man and
  Cybernetics (SMC)}, pages 21--26. IEEE, 2019.
\newblock \doi{10.1109/SMC.2019.8913901}.

\bibitem[Chalapathy and Chawla(2019)]{chal19}
Raghavendra Chalapathy and Sanjay Chawla.
\newblock Deep learning for anomaly detection: A survey.
\newblock \emph{arXiv preprint arXiv:1901.03407}, 2019.
\newblock URL \url{https://arxiv.org/pdf/1901.03407.pdf}.

\bibitem[Chen et~al.(2018)Chen, Song, Wainwright, and Jordan]{chen18}
Jianbo Chen, Le~Song, Martin Wainwright, and Michael Jordan.
\newblock Learning to explain: An information-theoretic perspective on model
  interpretation.
\newblock In Jennifer Dy and Andreas Krause, editors, \emph{Proceedings of the
  35th International Conference on Machine Learning}, volume~80 of
  \emph{Proceedings of Machine Learning Research}, pages 883--892. PMLR, 10--15
  Jul 2018.
\newblock URL \url{https://proceedings.mlr.press/v80/chen18j.html}.

\bibitem[Dau et~al.(2019)Dau, Bagnall, Kamgar, Yeh, Zhu, Gharghabi,
  Ratanamahatana, and Keogh]{dau19ucr}
Hoang~Anh Dau, Anthony Bagnall, Kaveh Kamgar, Chin-Chia~Michael Yeh, Yan Zhu,
  Shaghayegh Gharghabi, Chotirat~Ann Ratanamahatana, and Eamonn Keogh.
\newblock The ucr time series archive.
\newblock \emph{IEEE/CAA Journal of Automatica Sinica}, 6\penalty0
  (6):\penalty0 1293--1305, 2019.
\newblock \doi{10.1109/JAS.2019.1911747}.

\bibitem[Dempster et~al.(2020)Dempster, Petitjean, and Webb]{demp20}
Angus Dempster, Fran{\c{c}}ois Petitjean, and Geoffrey~I Webb.
\newblock {ROCKET}: exceptionally fast and accurate time series classification
  using random convolutional kernels.
\newblock \emph{Data Mining and Knowledge Discovery}, 34\penalty0 (5):\penalty0
  1454--1495, 2020.
\newblock \doi{10.1007/s10618-020-00701-z}.

\bibitem[Dempster et~al.(2021)Dempster, Schmidt, and Webb]{dem21}
Angus Dempster, Daniel~F. Schmidt, and Geoffrey~I. Webb.
\newblock \emph{{MiniRocket}: A Very Fast (Almost) Deterministic Transform for
  Time Series Classification}, page 248–257.
\newblock Association for Computing Machinery, New York, NY, USA, 2021.
\newblock ISBN 9781450383325.
\newblock \doi{10.1145/3447548.3467231}.

\bibitem[Deng et~al.(2013)Deng, Runger, Tuv, and Vladimir]{Deng13}
Houtao Deng, George Runger, Eugene Tuv, and Martyanov Vladimir.
\newblock A time series forest for classification and feature extraction.
\newblock \emph{Information Sciences}, 239:\penalty0 142--153, 2013.
\newblock ISSN 0020-0255.
\newblock \doi{10.1016/j.ins.2013.02.030}.

\bibitem[Fawaz et~al.(2020)Fawaz, Lucas, Forestier, Pelletier, Schmidt, Weber,
  Webb, Idoumghar, Muller, and Petitjean]{fawa20}
Hassan~Ismail Fawaz, Benjamin Lucas, Germain Forestier, Charlotte Pelletier,
  Daniel~F Schmidt, Jonathan Weber, Geoffrey~I Webb, Lhassane Idoumghar,
  Pierre-Alain Muller, and Fran{\c{c}}ois Petitjean.
\newblock {InceptionTime}: Finding alexnet for time series classification.
\newblock \emph{Data Mining and Knowledge Discovery}, 34\penalty0 (6):\penalty0
  1936--1962, 2020.
\newblock \doi{10.1007/s10618-020-00710-y}.

\bibitem[Guidotti et~al.(2018{\natexlab{a}})Guidotti, Monreale, Ruggieri,
  Pedreschi, Turini, and Giannotti]{guid18}
Riccardo Guidotti, Anna Monreale, Salvatore Ruggieri, Dino Pedreschi, Franco
  Turini, and Fosca Giannotti.
\newblock Local rule-based explanations of black box decision systems.
\newblock \emph{arXiv preprint arXiv:1805.10820}, 2018{\natexlab{a}}.
\newblock URL \url{https://arxiv.org/pdf/1805.10820.pdf}.

\bibitem[Guidotti et~al.(2018{\natexlab{b}})Guidotti, Monreale, Ruggieri,
  Turini, Giannotti, and Pedreschi]{guid19}
Riccardo Guidotti, Anna Monreale, Salvatore Ruggieri, Franco Turini, Fosca
  Giannotti, and Dino Pedreschi.
\newblock A survey of methods for explaining black box models.
\newblock \emph{ACM Comput. Surv.}, 51\penalty0 (5), aug 2018{\natexlab{b}}.
\newblock ISSN 0360-0300.
\newblock \doi{10.1145/3236009}.

\bibitem[Guo et~al.(2018)Guo, Mu, Xu, Su, Wang, and Xing]{guo18}
Wenbo Guo, Dongliang Mu, Jun Xu, Purui Su, Gang Wang, and Xinyu Xing.
\newblock Lemna: Explaining deep learning based security applications.
\newblock In \emph{Proceedings of the 2018 ACM SIGSAC Conference on Computer
  and Communications Security}, CCS '18, page 364–379, New York, NY, USA,
  2018. Association for Computing Machinery.
\newblock ISBN 9781450356930.
\newblock \doi{10.1145/3243734.3243792}.

\bibitem[He et~al.(2016)He, Zhang, Ren, and Sun]{he16}
Kaiming He, Xiangyu Zhang, Shaoqing Ren, and Jian Sun.
\newblock Deep residual learning for image recognition.
\newblock In \emph{Proceedings of the IEEE conference on computer vision and
  pattern recognition}, pages 770--778, 2016.
\newblock URL
  \url{https://www.cs.princeton.edu/courses/archive/spring16/cos598F/msra-deepnet.pdf}.

\bibitem[Hills et~al.(2014)Hills, Lines, Baranauskas, Mapp, and
  Bagnall]{hill14}
Jon Hills, Jason Lines, Edgaras Baranauskas, James Mapp, and Anthony Bagnall.
\newblock Classification of time series by shapelet transformation.
\newblock \emph{Data mining and knowledge discovery}, 28\penalty0 (4):\penalty0
  851--881, 2014.
\newblock \doi{10.1007/s10618-013-0322-1}.

\bibitem[Jacob et~al.(2021)Jacob, Song, Stiegler, Rad, Diao, and Tatbul]{jac21}
Vincent Jacob, Fei Song, Arnaud Stiegler, Bijan Rad, Yanlei Diao, and Nesime
  Tatbul.
\newblock {Exathlon: A Benchmark for Explainable Anomaly Detection over Time
  Series}.
\newblock \emph{{Proceedings of the VLDB Endowment (PVLDB)}}, July 2021.
\newblock URL \url{https://hal.archives-ouvertes.fr/hal-03381732}.

\bibitem[Jeyakumar et~al.(2019)Jeyakumar, Madani, Parandeh, Kulshreshtha, Zeng,
  and Yadav]{jeya19}
Vimalkumar Jeyakumar, Omid Madani, Ali Parandeh, Ashutosh Kulshreshtha, Weifei
  Zeng, and Navindra Yadav.
\newblock Explainit! -- a declarative root-cause analysis engine for time
  series data.
\newblock In \emph{Proceedings of the 2019 International Conference on
  Management of Data}, SIGMOD '19, page 333–348, New York, NY, USA, 2019.
  Association for Computing Machinery.
\newblock ISBN 9781450356435.
\newblock \doi{10.1145/3299869.3314048}.

\bibitem[Kate(2016)]{kate16}
Rohit~J Kate.
\newblock Using dynamic time warping distances as features for improved time
  series classification.
\newblock \emph{Data Mining and Knowledge Discovery}, 30\penalty0 (2):\penalty0
  283--312, 2016.
\newblock \doi{10.1007/s10618-015-0418-x}.

\bibitem[Keogh and Mueen(2017)]{keog17}
Eamonn Keogh and Abdullah Mueen.
\newblock \emph{Curse of Dimensionality}, pages 314--315.
\newblock Springer US, Boston, MA, 2017.
\newblock ISBN 978-1-4899-7687-1.
\newblock \doi{10.1007/978-1-4899-7687-1_192}.

\bibitem[Kiran et~al.(2018)Kiran, Thomas, and Parakkal]{kira17}
B.~Ravi Kiran, Dilip~Mathew Thomas, and Ranjith Parakkal.
\newblock An overview of deep learning based methods for unsupervised and
  semi-supervised anomaly detection in videos.
\newblock \emph{Journal of Imaging}, 4\penalty0 (2), 2018.
\newblock ISSN 2313-433X.
\newblock \doi{10.3390/jimaging4020036}.
\newblock URL \url{https://www.mdpi.com/2313-433X/4/2/36}.

\bibitem[Lapuschkin et~al.(2019)Lapuschkin, W{\"a}ldchen, Binder, Montavon,
  Samek, and M{\"u}ller]{lapu19}
Sebastian Lapuschkin, Stephan W{\"a}ldchen, Alexander Binder, Gr{\'e}goire
  Montavon, Wojciech Samek, and Klaus-Robert M{\"u}ller.
\newblock Unmasking clever hans predictors and assessing what machines really
  learn.
\newblock \emph{Nature Communications}, 10:\penalty0 1096, 2019.
\newblock URL \url{http://dx.doi.org/10.1038/s41467-019-08987-4}.

\bibitem[Large et~al.(2019)Large, Lines, and Bagnall]{larg19}
James Large, Jason Lines, and Anthony Bagnall.
\newblock A probabilistic classifier ensemble weighting scheme based on
  cross-validated accuracy estimates.
\newblock \emph{Data mining and knowledge discovery}, 33\penalty0 (6):\penalty0
  1674--1709, 2019.
\newblock \doi{10.1007/s10618-019-00638-y}.

\bibitem[Lines and Bagnall(2015)]{line15}
Jason Lines and Anthony Bagnall.
\newblock Time series classification with ensembles of elastic distance
  measures.
\newblock \emph{Data Mining and Knowledge Discovery}, 29\penalty0 (3):\penalty0
  565--592, 2015.

\bibitem[Lines et~al.(2018)Lines, Taylor, and Bagnall]{line18}
Jason Lines, Sarah Taylor, and Anthony Bagnall.
\newblock Time series classification with hive-cote: The hierarchical vote
  collective of transformation-based ensembles.
\newblock \emph{ACM Transactions on Knowledge Discovery from Data}, 12\penalty0
  (5), 2018.
\newblock \doi{10.1145/3182382}.

\bibitem[Liu et~al.(2008)Liu, Ting, and Zhou]{liu08}
Fei~Tony Liu, Kai~Ming Ting, and Zhi-Hua Zhou.
\newblock Isolation {F}orest.
\newblock In \emph{2008 Eighth IEEE International Conference on Data Mining},
  pages 413--422. IEEE, 2008.
\newblock \doi{10.1109/ICDM.2008.17}.

\bibitem[Liznerski et~al.(2021)Liznerski, Ruff, Vandermeulen, Franks, Kloft,
  and M{\"u}ller]{liz21}
Philipp Liznerski, Lukas Ruff, Robert~A Vandermeulen, Billy~Joe Franks, Marius
  Kloft, and Klaus-Robert M{\"u}ller.
\newblock Explainable deep one-class classification.
\newblock \emph{arXiv preprint arXiv:2007.01760}, 2021.
\newblock URL \url{https://openreview.net/pdf?id=A5VV3UyIQz}.

\bibitem[Lundberg and Lee(2017)]{lund17}
Scott~M Lundberg and Su-In Lee.
\newblock A unified approach to interpreting model predictions.
\newblock In \emph{Proceedings of the 31st international conference on neural
  information processing systems}, pages 4768--4777, 2017.
\newblock URL
  \url{https://proceedings.neurips.cc//paper/2017/file/8a20a8621978632d76c43dfd28b67767-Paper.pdf}.

\bibitem[Ma et~al.(2020)Ma, Yin, Zhang, Wang, Zheng, Jiang, Hu, Luo, Li, Qiu,
  et~al.]{ma20}
Minghua Ma, Zheng Yin, Shenglin Zhang, Sheng Wang, Christopher Zheng, Xinhao
  Jiang, Hanwen Hu, Cheng Luo, Yilin Li, Nengjun Qiu, et~al.
\newblock Diagnosing root causes of intermittent slow queries in cloud
  databases.
\newblock \emph{Proceedings of the VLDB Endowment}, 13\penalty0 (8):\penalty0
  1176--1189, 2020.
\newblock URL
  \url{http://nkcs.iops.ai/wp-content/uploads/2020/04/paper-VLDB20-iSQUAD.pdf}.

\bibitem[Maas et~al.(2013)Maas, Hannun, and Ng]{Maas13}
Andrew~L. Maas, Awni~Y. Hannun, and Andrew~Y. Ng.
\newblock Rectifier nonlinearities improve neural network acoustic models.
\newblock In \emph{in ICML Workshop on Deep Learning for Audio, Speech and
  Language Processing}, 2013.

\bibitem[Mathonsi and {van Zyl}(2022{\natexlab{a}})]{Mat2}
Thabang Mathonsi and Terence~L {van Zyl}.
\newblock Multivariate anomaly detection based on prediction intervals
  constructed using deep learning.
\newblock \emph{Neural Computing and Applications}, pages 1--15,
  2022{\natexlab{a}}.
\newblock \doi{10.1007/s00521-021-06697-x}.

\bibitem[Mathonsi and {van Zyl}(2022{\natexlab{b}})]{Mat3}
Thabang Mathonsi and Terence~L. {van Zyl}.
\newblock A statistics and deep learning hybrid method for multivariate time
  series forecasting and mortality modeling.
\newblock \emph{Forecasting}, 4\penalty0 (1):\penalty0 1--25,
  2022{\natexlab{b}}.
\newblock ISSN 2571-9394.
\newblock \doi{10.3390/forecast4010001}.

\bibitem[Middlehurst et~al.(2019)Middlehurst, Vickers, and Bagnall]{midd19}
Matthew Middlehurst, William Vickers, and Anthony Bagnall.
\newblock Scalable dictionary classifiers for time series classification.
\newblock In \emph{International Conference on Intelligent Data Engineering and
  Automated Learning}, pages 11--19. Springer, 2019.
\newblock \doi{10.1007/978-3-030-33607-3_2}.

\bibitem[Mitchell et~al.(2019)Mitchell, Wu, Zaldivar, Barnes, Vasserman,
  Hutchinson, Spitzer, Raji, and Gebru]{mitc19}
Margaret Mitchell, Simone Wu, Andrew Zaldivar, Parker Barnes, Lucy Vasserman,
  Ben Hutchinson, Elena Spitzer, Inioluwa~Deborah Raji, and Timnit Gebru.
\newblock Model cards for model reporting.
\newblock In \emph{Proceedings of the Conference on Fairness, Accountability,
  and Transparency}, FAT* '19, page 220–229, New York, NY, USA, 2019.
  Association for Computing Machinery.
\newblock ISBN 9781450361255.
\newblock \doi{10.1145/3287560.3287596}.

\bibitem[Munir et~al.(2019)Munir, Chattha, Dengel, and Ahmed]{muni19}
Mohsin Munir, Muhammad~Ali Chattha, Andreas Dengel, and Sheraz Ahmed.
\newblock A comparative analysis of traditional and deep learning-based anomaly
  detection methods for streaming data.
\newblock In M.~Arif Wani, Taghi~M. Khoshgoftaar, Dingding Wang, Huanjing Wang,
  and Naeem Seliya, editors, \emph{ICMLA}, pages 561--566. IEEE, 2019.
\newblock \doi{10.1109/ICMLA.2019.00105}.

\bibitem[Neamtu et~al.(2018)Neamtu, Ahsan, Rundensteiner, Sarkozy, Keogh, Dau,
  Nguyen, and Lovering]{neam18}
Rodica Neamtu, Ramoza Ahsan, Elke~A. Rundensteiner, Gabor Sarkozy, Eamonn
  Keogh, Hoang~Anh Dau, Cuong Nguyen, and Charles Lovering.
\newblock Generalized dynamic time warping: Unleashing the warping power hidden
  in point-wise distances.
\newblock In \emph{2018 IEEE 34th International Conference on Data Engineering
  (ICDE)}, pages 521--532, 2018.
\newblock \doi{10.1109/ICDE.2018.00054}.

\bibitem[Nguyen et~al.(2019)Nguyen, Lim, Divakaran, Low, and Chan]{nguy19}
Quoc~Phong Nguyen, Kar~Wai Lim, Dinil~Mon Divakaran, Kian~Hsiang Low, and
  Mun~Choon Chan.
\newblock Gee: A gradient-based explainable variational autoencoder for network
  anomaly detection.
\newblock In \emph{2019 IEEE Conference on Communications and Network Security
  (CNS)}, pages 91--99. IEEE, 2019.
\newblock \doi{10.1109/CNS.2019.8802833}.

\bibitem[Pang et~al.(2020)Pang, Shen, Cao, and van~den Hengel]{pang20}
Guansong Pang, Chunhua Shen, Longbing Cao, and Anton van~den Hengel.
\newblock Deep learning for anomaly detection: {A} review.
\newblock \emph{CoRR}, abs/2007.02500, 2020.
\newblock URL \url{https://arxiv.org/abs/2007.02500}.

\bibitem[Pang et~al.(2021)Pang, Ding, Shen, and Hengel]{pang21}
Guansong Pang, Choubo Ding, Chunhua Shen, and Anton van~den Hengel.
\newblock Explainable deep few-shot anomaly detection with deviation networks.
\newblock \emph{arXiv preprint arXiv:2108.00462}, 2021.
\newblock URL \url{https://arxiv.org/pdf/2108.00462.pdf}.

\bibitem[Plumb et~al.(2018)Plumb, Molitor, and Talwalkar]{plum18}
Gregory Plumb, Denali Molitor, and Ameet~S Talwalkar.
\newblock Model agnostic supervised local explanations.
\newblock In S.~Bengio, H.~Wallach, H.~Larochelle, K.~Grauman, N.~Cesa-Bianchi,
  and R.~Garnett, editors, \emph{Advances in Neural Information Processing
  Systems}, volume~31. Curran Associates, Inc., 2018.
\newblock URL
  \url{https://proceedings.neurips.cc/paper/2018/file/b495ce63ede0f4efc9eec62cb947c162-Paper.pdf}.

\bibitem[Rad et~al.(2021)Rad, Song, Jacob, and Diao]{rad21}
Bijan Rad, Fei Song, Vincent Jacob, and Yanlei Diao.
\newblock Explainable anomaly detection on high-dimensional time series data.
\newblock In \emph{Proceedings of the 15th ACM International Conference on
  Distributed and Event-Based Systems}, DEBS '21, page 2–14, New York, NY,
  USA, 2021. Association for Computing Machinery.
\newblock ISBN 9781450385558.
\newblock \doi{10.1145/3465480.3468292}.

\bibitem[Ribeiro et~al.(2016)Ribeiro, Singh, and Guestrin]{ribe16}
Marco~Tulio Ribeiro, Sameer Singh, and Carlos Guestrin.
\newblock "why should i trust you?": Explaining the predictions of any
  classifier.
\newblock In \emph{Proceedings of the 22nd ACM SIGKDD International Conference
  on Knowledge Discovery and Data Mining}, KDD '16, page 1135–1144, New York,
  NY, USA, 2016. Association for Computing Machinery.
\newblock ISBN 9781450342322.
\newblock \doi{10.1145/2939672.2939778}.

\bibitem[Ribeiro et~al.(2018)Ribeiro, Singh, and Guestrin]{Ribe18}
Marco~Tulio Ribeiro, Sameer Singh, and Carlos Guestrin.
\newblock Anchors: High-precision model-agnostic explanations.
\newblock \emph{Proceedings of the {AAAI} Conference on Artificial
  Intelligence}, 32\penalty0 (1), Apr. 2018.
\newblock URL \url{https://ojs.aaai.org/index.php/AAAI/article/view/11491}.

\bibitem[Ruff et~al.(2021)Ruff, Kauffmann, Vandermeulen, Montavon, Samek,
  Kloft, Dietterich, and M{\"u}ller]{ruff21}
Lukas Ruff, Jacob~R Kauffmann, Robert~A Vandermeulen, Gr{\'e}goire Montavon,
  Wojciech Samek, Marius Kloft, Thomas~G Dietterich, and Klaus-Robert
  M{\"u}ller.
\newblock A unifying review of deep and shallow anomaly detection.
\newblock \emph{Proceedings of the IEEE}, 109\penalty0 (5):\penalty0 756--795,
  2021.
\newblock \doi{10.1109/JPROC.2021.3052449}.

\bibitem[Ruiz et~al.(2021)Ruiz, Flynn, Large, Middlehurst, and Bagnall]{ruiz21}
Alejandro~Pasos Ruiz, Michael Flynn, James Large, Matthew Middlehurst, and
  Anthony Bagnall.
\newblock The great multivariate time series classification bake off: a review
  and experimental evaluation of recent algorithmic advances.
\newblock \emph{Data Mining and Knowledge Discovery}, 35\penalty0 (2):\penalty0
  401--449, 2021.
\newblock URL
  \url{https://ueaeprints.uea.ac.uk/id/eprint/77815/7/Ruiz2020_Article_TheGreatMultivariateTimeSeries.pdf}.

\bibitem[Sainath et~al.(2013)Sainath, Kingsbury, Mohamed, Dahl, Saon, Soltau,
  Beran, Aravkin, and Ramabhadran]{sain13}
Tara~N. Sainath, Brian Kingsbury, Abdel-rahman Mohamed, George~E. Dahl, George
  Saon, Hagen Soltau, Tomas Beran, Aleksandr~Y. Aravkin, and Bhuvana
  Ramabhadran.
\newblock Improvements to deep convolutional neural networks for lvcsr.
\newblock In \emph{2013 IEEE Workshop on Automatic Speech Recognition and
  Understanding}, pages 315--320, 2013.
\newblock \doi{10.1109/ASRU.2013.6707749}.

\bibitem[Selvaraju et~al.(2017)Selvaraju, Cogswell, Das, Vedantam, Parikh, and
  Batra]{selv17}
Ramprasaath~R Selvaraju, Michael Cogswell, Abhishek Das, Ramakrishna Vedantam,
  Devi Parikh, and Dhruv Batra.
\newblock Grad-cam: Visual explanations from deep networks via gradient-based
  localization.
\newblock In \emph{Proceedings of the IEEE international conference on computer
  vision}, pages 618--626, 2017.
\newblock URL
  \url{https://openaccess.thecvf.com/content_ICCV_2017/papers/Selvaraju_Grad-CAM_Visual_Explanations_ICCV_2017_paper.pdf}.

\bibitem[Serradilla et~al.(2021)Serradilla, Zugasti, Ramirez~de Okariz,
  Rodriguez, and Zurutuza]{sera21}
Oscar Serradilla, Ekhi Zugasti, Julian Ramirez~de Okariz, Jon Rodriguez, and
  Urko Zurutuza.
\newblock Adaptable and explainable predictive maintenance: Semi-supervised
  deep learning for anomaly detection and diagnosis in press machine data.
\newblock \emph{Applied Sciences}, 11\penalty0 (16), 2021.
\newblock ISSN 2076-3417.
\newblock \doi{10.3390/app11167376}.

\bibitem[Shokoohi-Yekta et~al.(2017)Shokoohi-Yekta, Hu, Jin, Wang, and
  Keogh]{shok17}
Mohammad Shokoohi-Yekta, Bing Hu, Hongxia Jin, Jun Wang, and Eamonn Keogh.
\newblock Generalizing dtw to the multi-dimensional case requires an adaptive
  approach.
\newblock \emph{Data mining and knowledge discovery}, 31\penalty0 (1):\penalty0
  1--31, 2017.
\newblock \doi{10.1007/s10618-016-0455-0}.

\bibitem[Shrikumar et~al.(2017)Shrikumar, Greenside, and Kundaje]{shri17}
Avanti Shrikumar, Peyton Greenside, and Anshul Kundaje.
\newblock Learning important features through propagating activation
  differences.
\newblock In \emph{International Conference on Machine Learning}, pages
  3145--3153. PMLR, 2017.
\newblock URL
  \url{http://proceedings.mlr.press/v70/shrikumar17a/shrikumar17a.pdf}.

\bibitem[Siddiqui et~al.(2019)Siddiqui, Mercier, Munir, Dengel, and
  Ahmed]{sidd19}
Shoaib~Ahmed Siddiqui, Dominique Mercier, Mohsin Munir, Andreas Dengel, and
  Sheraz Ahmed.
\newblock Tsviz: Demystification of deep learning models for time-series
  analysis.
\newblock \emph{IEEE Access}, 7:\penalty0 67027--67040, 2019.
\newblock \doi{10.1109/ACCESS.2019.2912823}.

\bibitem[Simonyan et~al.(2013)Simonyan, Vedaldi, and Zisserman]{simo13}
Karen Simonyan, Andrea Vedaldi, and Andrew Zisserman.
\newblock Deep inside convolutional networks: Visualising image classification
  models and saliency maps.
\newblock \emph{arXiv preprint arXiv:1312.6034}, 2013.
\newblock URL \url{https://arxiv.org/pdf/1312.6034.pdf}.

\bibitem[Sundararajan et~al.(2017)Sundararajan, Taly, and Yan]{sund17}
Mukund Sundararajan, Ankur Taly, and Qiqi Yan.
\newblock Axiomatic attribution for deep networks.
\newblock In Doina Precup and Yee~Whye Teh, editors, \emph{Proceedings of the
  34th International Conference on Machine Learning}, volume~70 of
  \emph{Proceedings of Machine Learning Research}, pages 3319--3328. PMLR,
  06--11 Aug 2017.
\newblock URL \url{https://proceedings.mlr.press/v70/sundararajan17a.html}.

\bibitem[Szegedy et~al.(2015)Szegedy, Liu, Jia, Sermanet, Reed, Anguelov,
  Erhan, Vanhoucke, and Rabinovich]{szeg15}
Christian Szegedy, Wei Liu, Yangqing Jia, Pierre Sermanet, Scott Reed, Dragomir
  Anguelov, Dumitru Erhan, Vincent Vanhoucke, and Andrew Rabinovich.
\newblock Going deeper with convolutions.
\newblock In \emph{Proceedings of the IEEE conference on computer vision and
  pattern recognition}, pages 1--9, 2015.
\newblock \doi{10.1109/CVPR.2015.7298594}.

\bibitem[Tjoa and Guan(2021)]{tjoa20}
Erico Tjoa and Cuntai Guan.
\newblock A survey on explainable artificial intelligence (xai): Toward medical
  xai.
\newblock \emph{IEEE Transactions on Neural Networks and Learning Systems},
  32\penalty0 (11):\penalty0 4793--4813, 2021.
\newblock \doi{10.1109/TNNLS.2020.3027314}.

\bibitem[Veerappa et~al.(2021)Veerappa, Anneken, and Burkart]{veer21}
Manjunatha Veerappa, Mathias Anneken, and Nadia Burkart.
\newblock Evaluation of interpretable association rule mining methods on
  time-series in the maritime domain.
\newblock In \emph{International Conference on Pattern Recognition}, pages
  204--218. Springer, 2021.
\newblock \doi{10.1007/978-3-030-68796-0_15}.

\bibitem[Wang et~al.(2017)Wang, Yan, and Oates]{wang17}
Zhiguang Wang, Weizhong Yan, and Tim Oates.
\newblock Time series classification from scratch with deep neural networks: A
  strong baseline.
\newblock In \emph{2017 International joint conference on neural networks
  (IJCNN)}, pages 1578--1585. IEEE, 2017.
\newblock \doi{10.1109/IJCNN.2017.7966039}.

\bibitem[Westerski et~al.(2021)Westerski, Kanagasabai, Shaham, Narayanan, Wong,
  and Singh]{west21}
Adam Westerski, Rajaraman Kanagasabai, Eran Shaham, Amudha Narayanan, Jiayu
  Wong, and Manjeet Singh.
\newblock Explainable anomaly detection for procurement fraud
  identification—lessons from practical deployments.
\newblock \emph{International Transactions in Operational Research},
  28\penalty0 (6):\penalty0 3276--3302, 2021.
\newblock \doi{10.1111/itor.12968}.

\bibitem[Wu et~al.(2021)Wu, Shao, Tunc, Satam, and Hariri]{wu21}
Chongke Wu, Sicong Shao, Cihan Tunc, Pratik Satam, and Salim Hariri.
\newblock An explainable and efficient deep learning framework for video
  anomaly detection.
\newblock \emph{Cluster Computing}, 2021.
\newblock \doi{10.1007/s10586-021-03439-5}.

\bibitem[Yang et~al.(2017)Yang, Rudin, and Seltzer]{hong17}
Hongyu Yang, Cynthia Rudin, and Margo Seltzer.
\newblock Scalable {B}ayesian rule lists.
\newblock In Doina Precup and Yee~Whye Teh, editors, \emph{Proceedings of the
  34th International Conference on Machine Learning}, volume~70 of
  \emph{Proceedings of Machine Learning Research}, pages 3921--3930. PMLR,
  06--11 Aug 2017.
\newblock URL \url{https://proceedings.mlr.press/v70/yang17h.html}.

\bibitem[Zhang et~al.(2020)Zhang, Gao, Lin, and Lu]{zhan20}
Xuchao Zhang, Yifeng Gao, Jessica Lin, and Chang-Tien Lu.
\newblock {TapNet}: Multivariate time series classification with attentional
  prototypical network.
\newblock In \emph{Proceedings of the AAAI Conference on Artificial
  Intelligence}, volume~34, pages 6845--6852, 2020.
\newblock \doi{10.1609/aaai.v34i04.6165}.

\bibitem[Zheng et~al.(2021{\natexlab{a}})Zheng, Xu, Trinh, Wu, Huang,
  Sivaranjani, Liu, and Xie]{Zhen21}
Xiangtian Zheng, Nan Xu, Loc Trinh, Dongqi Wu, Tong Huang, S~Sivaranjani, Yan
  Liu, and Le~Xie.
\newblock Psml: A multi-scale time-series dataset for machine learning in
  decarbonized energy grids.
\newblock \emph{arXiv preprint arXiv:2110.06324}, 2021{\natexlab{a}}.
\newblock URL \url{https://arxiv.org/pdf/2110.06324.pdf}.

\bibitem[Zheng et~al.(2021{\natexlab{b}})Zheng, Xu, Wu, Trinh, Huang,
  Sivaranjani, Liu, and Xie]{psml}
Xiangtian Zheng, Nan Xu, Dongqi Wu, Loc Trinh, Tong Huang, S~Sivaranjani, Yan
  Liu, and Le~Xie.
\newblock {PSML: A Multi-scale Time-series Dataset for Machine Learning in
  Decarbonized Energy Grids (Dataset)}, August 2021{\natexlab{b}}.

\bibitem[Zheng et~al.(2014)Zheng, Liu, Chen, Ge, and Zhao]{zhen14}
Yi~Zheng, Qi~Liu, Enhong Chen, Yong Ge, and J~Leon Zhao.
\newblock Time series classification using multi-channels deep convolutional
  neural networks.
\newblock In \emph{International conference on web-age information management},
  pages 298--310. Springer, 2014.
\newblock \doi{10.1007/978-3-319-08010-9_33}.

\end{thebibliography}
\bibliographystyle{plainnat} 

\end{document}